\documentclass[11pt]{article}
\usepackage{float}
\restylefloat{table}
\usepackage{caption}
\usepackage{coling}
\usepackage{enumerate} 
\usepackage{times}
\usepackage{latexsym}

\usepackage[T1]{fontenc}

\usepackage[utf8]{inputenc}

\usepackage{microtype}

\usepackage{inconsolata}

\usepackage{graphicx}

\title{Fotheidil: an Automatic Transcription System for the Irish Language}

\author{
  \textbf{Liam Lonergan\textsuperscript{1}},
  \textbf{Ibon Saratxaga\textsuperscript{2}},
  \textbf{John Sloan\textsuperscript{1}},
  \textbf{Oscar Maharog \textsuperscript{1}},
\\
  \textbf{Mengjie Qian\textsuperscript{3}},
  \textbf{Neasa Ní Chiaráin\textsuperscript{1}},
  \textbf{Christer Gobl\textsuperscript{1}},
  \textbf{Ailbhe Ní Chasaide \textsuperscript{1}}
\\\\
  \textsuperscript{1}Phonetics and Speech Laboratory, Trinity College Dublin,\\
  \textsuperscript{2}HiTZ Basque Center for Language Technology, AhoLab,\\ University of the Basque Country UPV/EHU,
  \\
  \textsuperscript{3}Department of Engineering, University of Cambridge
\\
  \small{
  \textsuperscript{1}\{llonerga, sloanjo,mahargbo, neasa.nichiarain, cegobl, anichsid\}@tcd.ie, \textsuperscript{2}ibon.saratxaga@ehu.eus, \textsuperscript{3}mq227@cam.ac.uk
  }
}

\begin{document}
\maketitle
\begin{abstract}
This paper sets out the first web-based transcription system for the Irish language - Fotheidil, a system that utilises speech-related AI technologies as part of the ABAIR initiative. The system includes both off-the-shelf pre-trained voice activity detection and speaker diarisation models and models trained specifically for Irish automatic speech recognition and capitalisation and punctuation restoration. Semi-supervised learning is explored to improve the acoustic model of a modular TDNN-HMM ASR system, yielding substantial improvements for out-of-domain test sets and dialects that are underrepresented in the supervised training set. A novel approach to capitalisation and punctuation restoration involving sequence-to-sequence models is compared with the conventional approach using a classification model. Experimental results show here also substantial improvements in performance. The system will be made freely available for public use, and represents an important resource to researchers and others who transcribe Irish language materials. Human-corrected transcriptions will be collected and included in the training dataset as the system is used, which should lead to incremental improvements to the ASR model in a cyclical, community-driven fashion. \end{abstract}

\section{Introduction}
Artificial intelligence (AI) has become a pervasive part of today's world. While AI undoubtedly brings many benefits, these benefits are felt primarily by speaker communities of the world's major languages. Speakers of minority languages have not been adequately serviced with technology that works for them and is appropriate for their needs. 

Automatic speech recognition (ASR), the process of automatically transcribing speech into text, is a prime example of this disparity. While modern systems for English or Chinese approximate, or even improve upon, the performance of human transcription, for most languages Speech-to-Text does not exist. One of the largest barriers to developing ASR systems for minority languages is a lack of large, transcribed speech corpora. Recently, approaches leveraging large unlabelled speech corpora, such as semi-supervised learning \cite{zhang2020pushing,radford2023robust} and self-supervised learning \cite{baevski2020wav2vec} have achieved state-of-the-art performance for common ASR benchmarks, and were beneficial for low-resource languages \cite{dehaven2022improving}.

Speech-to-Text integrated technologies like automatic closed captioning on platforms such as YouTube and TikTok have been widely adopted by users of major languages. However, lesser-resourced languages are not included in such services. In light of this gap, we present Fotheidil\footnote{\url{https://fotheidil.abair.ie}} - a freely available web-based transcription system for the Irish language that utilises various speech-related AI components to transcribe long audio and video files. The structure of the paper is as follows: Section 2 outlines relevant background information; Section 3 details the system Interface; and section 4 describes the transcription pipeline, and the experiments carried out to improve Irish ASR performance using semi-supervised learning (SSL), as well as experiments carried out to train a Capitalisation and Punctuation Restoration (C\&PR) model, which improves the legibility of the ASR outputs for the end-user; and finally, Section 5 contains the discussion and conclusions.

\section{Background}

\subsection{Irish language}
Irish, a Goidelic or Gaelic language, is a member of the Celtic branch of the Indo-European language family. Today, the Gaelic languages are spoken in small communities scattered mostly along the western seaboard of Ireland and the western islands of Scotland. The almost extinct Manx, which is currently being revived, is also a Goidelic language and is spoken on the Isle of Man. 

The Irish language is highly inflected and has a complex phonological system. The language is diverse in its dialects and accents, with three regional dialects of Ulster (Ul), Connaught (Co) and Munster (Mu) and further sub-dialects, as well as the accents of non-native speakers i.e., learners and new speakers (Nn). The dialects vary significantly in terms of pronunciation, vocabulary and grammar and the phonology and syntactic structure of non-native speakers can often approximate that of English. Speaker variety is used here to describe the dialect or accent of a speaker.


\subsection{ABAIR}

The ABAIR initiative has been developing speech technology and applications to close the technology gap for the Irish language. Synthetic voices for the 3 major dialects of the languages of Ulster (Ul), Connacht (Co) and Munster (Mu) have been developed, with plans to expand this to further sub-dialects. Additionally, speech recognition systems for Irish have been developed with a sociolinguistic focus, by ensuring that we have adequate coverage of the different varieties of the language where possible and by evaluating our systems for their performance on speakers of different varieties.

\subsection{Automatic Speech Recognition}
ASR, the task of converting speech into text, has seen significant progress in recent years, due to advances in deep learning, access to hardware such as graphical processing units (GPU) and the increasing use of very large speech corpora. There are two conventional approaches to ASR - the traditional, modular approach, where the speech-to-text task is broken into the distinct components of acoustic modeling, pronunciation lexicon, and language modeling. The sub-modules are modeled independently and then combined in a decoding graph as a weighted finite-state transducer. In contrast, End-to-End (E2E) systems handle the entire speech-to-text task within a single model, directly learning the mapping from audio to text without the need for separate modules, offering a more streamlined but data-intensive approach. While E2E systems have surpassed modular systems in most performance benchmarks for ASR, the need to use large training corpora makes them less suitable for low-resource languages \cite{lonergan24_odyssey}. 

\subsubsection{Semi-supervised learning}
One of the most significant bottlenecks to the development of speech recognition for an under-resourced language is the availability of transcribed audio material to train an ASR system in a supervised manner. However, untranscribed speech is more readily available, due to the increasing proliferation of audio and video materials on the internet. Semi-supervised learning (SSL) is a paradigm that seeks to incorporate large unlabelled datasets in the learning framework to reduce the reliance on a large amount of labelled data. Among various SSL techniques applied to ASR, Noisy Student Teacher training (NST) has gained significant attention, achieving state-of-the-art performances across various datasets \cite{zhang2020pushing,park2020improved}. Moreover, it has improved performance in code-switching ASR \cite{xi2024semi} and in low-resource ASR \cite{li2024improving}.

\textbf{Noisy Student Teacher training}: in NST training, a teacher model is trained with the available labelled data. This model is used to generate pseudo-labels for the unlabelled dataset, which is combined with the labelled data to create a new training set for the student model. Noise is introduced to the new training set, forcing the student model to learn to reproduce the teacher model's outputs under noisy conditions, steering the model to learn more robust features that may better match the variability of real use.

\subsection{Capitalisation and Punctuation Restoration}
The output of the ASR system consists of raw text, using just lowercase characters, without punctuation symbols, acronyms or digits. This format is not very suitable in terms of readability and thus an additional processing is needed to restore proper capitalization and punctuation. Typically, capitalisation and punctuation restoration (C\&PR) systems are word-level classifiers which implement either two separate classifiers for capitalisation and for punctuation, or a joint one. A review on these systems can be found in \citet{10.1007/s10462-021-10051-x}.

In the case of Irish language, while punctuation rules are analogous to other western languages, capitalisation has its particularities, due to intial mutation, which is indicated orthographically by attaching different particles at the beginning of a word. These particles are one or two letters and are always written in lowercase, while the word keeps its original capitalisation: \textit{i nGaeilge} (in Irish), \textit{ón bhFrainc} (from France). 

These specific cases are not covered by word level capitalisation models for other languages, which usually have just two classes indicating if the initial letter of the word should be lower or uppercased. There are also character level capitalisation systems, but they perform worse and they are not so common.

Besides capitalisation and punctuation, the readability of the text is improved if numbers are written in digits instead of their textual form and acronyms are written in their condensed form instead of as they have been uttered by the speaker. Additionally, the use of specific symbols like percentage, currency, ordinal markers is also desirable. These fully formatted texts are often referred to as rich transcriptions.

\section{Interface}

\begin{figure}[ht]
    \centering
    \includegraphics[width=0.48\textwidth]{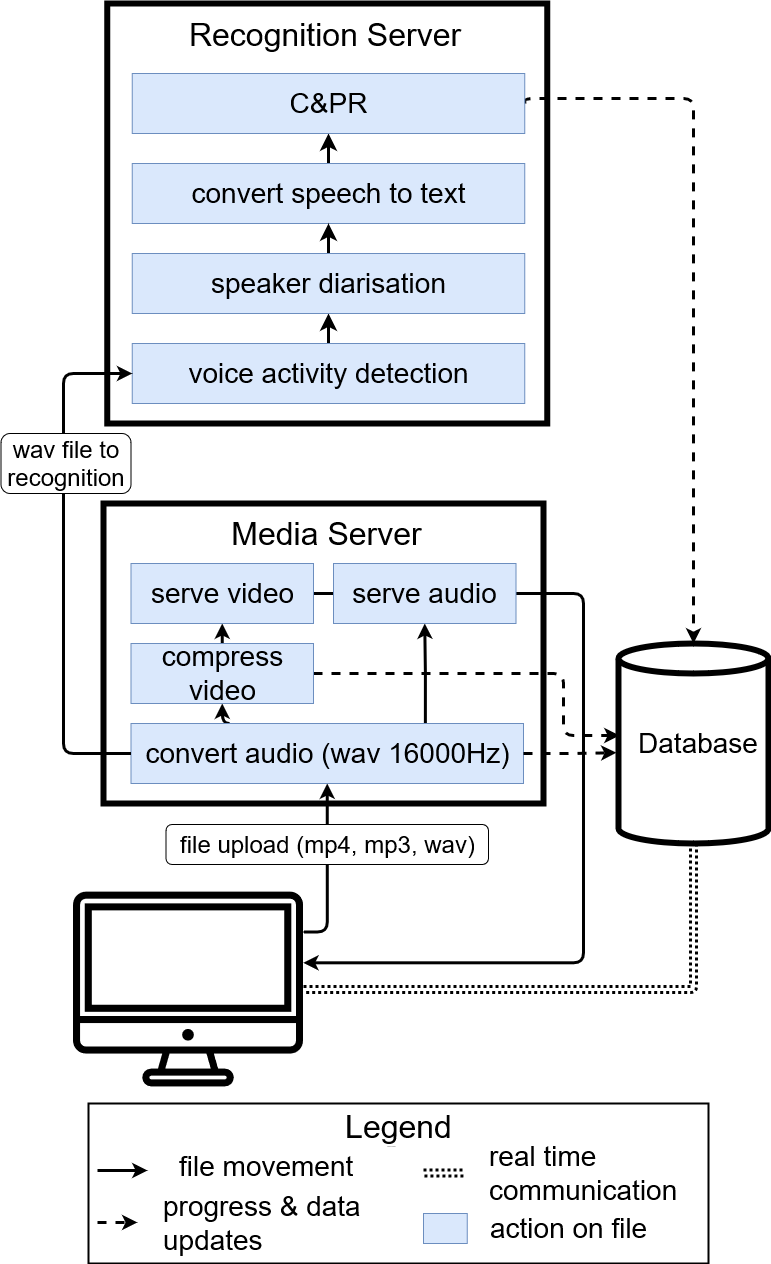} 
    \caption{Infrastructure Diagram}
    \label{fig:infrastructure}
\end{figure}

User experience (UX) has been a key concern for the development of the Fotheidil interface. Users of the platform are likely to experience significant wait times while the files they uploaded undergo processing and recognition. These wait times have been shown to exhibit a negative logarithmic relationship with user satisfaction \cite{egger2012waiting, reichl2010logarithmic}. Mitigating the negative effects of long loading times in UX is typically tackled in two ways: speeding up processing; and reducing the user’s frustration or perception of wait times. Increasing the processing speed is largely dependent on available hardware but can be aided by efficient infrastructure design. Reducing users’ perception of waiting times can be achieved on the front end through effective use of loading visualisations \cite{kim2017effect}. A description of both is provided below, with reference to Figure~\ref{fig:infrastructure}. 

\subsection{Backend}
Three main back-end functionalities are hosted on separate Virtual Machines (VMs) to avoid competition for CPU resources. Media processing is carried out on one VM, voice activity detection, speaker diarisation and ASR on another, with database storage on a third. When a user uploads a media file, it is first directed to the Media Server where the audio is stripped/converted to a wav file with a 16000 Hz sampling rate and video compression takes place if necessary. The converted wav file is then sent to the Recognition VM where speaker diarisation and recognition models are hosted. Updates on the progress for each of the processes with potentially long wait times are continuously stored on the database.

\begin{figure*}[ht]
   \centering
    \includegraphics[width=0.9\textwidth]{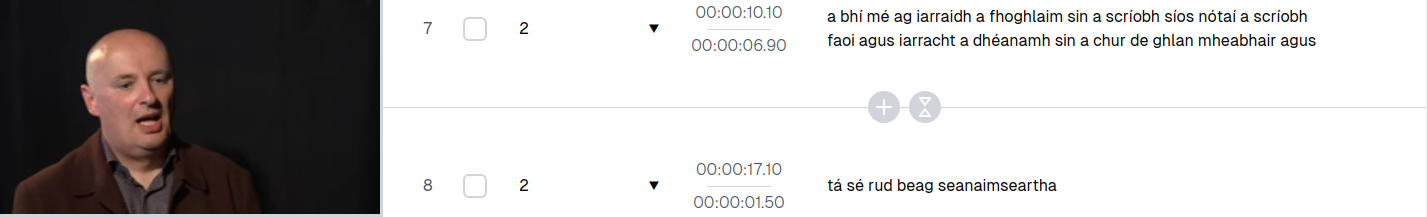} 
    \caption{Main User Interface}
    \label{fig:interface}
\end{figure*}


Real Time Communication (RTC) between the front end and the database enables the user to view progress for each of these back-end processing steps. The main dashboard for interacting with, and editting, the processed data is shown in Figure~\ref{fig:interface}. Users are able to edit the text, times and and speaker as well as download the output in pdf, docx or srt format.

\section{Transcription Pipeline and Experiments}
The transcription pipeline is a multi-step process, which brings together different systems to transcribe long audio files into text.
The choice of the models used in some of these steps is made by weighing up their performance and efficiency. 

The process is as follows:

\begin{enumerate}[i.] 
\setlength{\itemsep}{0 pt} 
    \item Upload audio or video file.
    \item Extract or convert audio to 16kHz mono wav.
    \item Voice activity detection to create segments.
    \item Speaker diarisation to assign speaker labels within speech segments.
    \item Continuous segments of same speaker are joined.
    \item Segments are decoded with ASR system.
    \item ASR output is enhanced using a C\&PR model.

\end{enumerate}

While there are better performing alternatives that make use of GPUs, our web-based service is limited to CPU usage only. The voice activity detection and speaker diarisation systems are off-the-shelf, pretrained models and are detailed briefly in Sections \ref{VAD} and \ref{SPK}, while the ASR and Punctuation and Capitalisaion models, which have been trained specifically for Irish, and these experiments are described in Sections \ref{ASR} and \ref{C&P}. 

\subsection{Voice Activity Detection} \label{VAD}
Voice activity detection is the process of finding speech segments within an audio file. The voice activity detection (VAD) module used is Silero-VAD {\cite{SileroVAD}} - a robust, lightweight, pre-trained model. While PyAnnote is the conventional off-the-shelf choice for both VAD and speaker diarisation, it requires a GPU to be used efficiently. Silero-VAD offers a CPU only alternative with competitive VAD performance. 

\subsection{Speaker Diarisation} \label{SPK}
The goal of speaker diarisation is to assign a speaker label to each speech segment. A pretrained Kaldi-based speaker diarisation x-vector \cite{snyder2018x} model\footnote{\url{https://kaldi-asr.org/models/m8}} that is trained using augmented VoxCeleb1 and VoxCeleb2 datasets is used as part of this pipeline. The model has a  reported EER performance of 3.7\% on the Speakers in the Wild speaker identification test set.

\subsection{Automatic Speech Recognition} \label{ASR}
Modular ASR approaches are often more suited to low-resource domains, as they do not require the same amount of data as E2E approaches. Additionally, modular systems are optimised to run efficiently on the CPU. Therefore, Kaldi-based DNN-HMM ASR models are used in our system and in the following experiments. 

A baseline supervised model M\textsubscript{0} is trained using the supervised training set of ~398h as described in Section \ref{Data} and detailed in Table \ref{tab:supervised_training_set} in the Appendix. To explore the usefulness of SSL for Irish ASR, a version of the approach outlined in \citet{manohar2018semi}, modified to include the noising element of NST, is tested in these experiments. The teacher model M\textsubscript{0} is used to decode the unlabelled set of 3230h in an undeterminised fashion, preserving the full decoding lattices. These lattices are rescored using a large n-gram language model (LM) and the best path through the rescored lattices is found. The best paths are taken as pseudo-labels and are combined with the supervised training set to create a semi-supervised training set. The student model M\textsubscript{1}  is trained with the semi-supervised training set using the noising technique Spectral Augment \cite{park2019specaugment}.

\subsubsection{Data} \label{Data}
The supervised acoustic training set comprises various datasets, as described in Table \ref{tab:supervised_training_set} in the Appendix with a breakdown of duration by speaker variety. Recordings used for ABAIR synthetic voices of the three dialects are used, totaling 41.4h (Syn). MíleGlór (MG) is an initiative for recording Irish speakers in the field and online using dialect-specific prompts, and a portion of 17.3h of this set is used. Additionally, two spontaneous speech (SS) corpora are combined, the large Corpas na Cainte Beo and the smaller Comhrá, totaling 259.7h. Audiobooks consisting of both professional and home recordings make up 36.6h. Caint Chonamara\footnote{\url{https://www.sksk.de/index.php/de/veroeffentlichungen-2/materialien/33-caint-chonamara}} (CCh), is a collection of conversations that was recorded in the Conamara area in 1964, representing rich dialectal speech of the Co dialect. Báiliúchán Béaloidis Árann (BBhÁ) is a folklore collection of conversational speech from the Aran Islands\footnote{\url{https://bba.duchas.ie/en/about/bba}}. Datasets SS, AB, CCh and BBhÁ were aligned using the alignment protocol set out in \hbox{\citet{lonergan24_odyssey}}.

The unsupervised acoustic data consists of broadcast recordings from four Irish language radio shows featured on Raidió na Gaeltachta: Barrscéalta, which mainly features speakers of Ul Irish; Adhmhaidin, which primarily contains speakers of the Co dialect; An Saol Ó Dheas which largely features Mu speakers; and Nuacht a hAon, which has a mix of dialects. These recordings are downloadable in MP3 format from Raidió na Gaeltachta's podcast page. Silero-VAD, as described in Section \ref{VAD}, is used to find speech chunks for decoding and resulted in 3230h. A breakdown in duration by radio show is provided in Table \ref{table:unsupervised_speech_corpus} in the Appendix.

Five test sets are used to evaluate the system and details for these sets are given in Table \ref{tab:test_set_duration_num_utts} in the Appendix. The first two are portions of MG and SS taken from the training set corpora, ensuring no data leakage, and can be considered as in-domain tests. These sets are 10.2h and 28.2h in length respectively and their speaker variety breakdown in duration is detailed in Table \ref{tab:test_set_by_speaker_var} in the Appendix. Two additional out-of-domain test sets are the Irish test portions of CommonVoice (CV) \cite{ardila-etal-2020-common} and Fleurs-R (FL) \cite{ma2024fleurs} datasets.  The quality of these datasets is markedly poor. Both datasets feature predominately non-native (Nn) speakers, and the texts for FL seem to be machine-translated English texts, which contain many foreign proper nouns. However, as they are publicly available and out-of-domain, we have included them here. They are 0.6h and 2.2h long respectively. Finally, 10 minutes each from the four radio shows (0.7h) from which the unsupervised dataset is created, were hand-labelled by the authors and are used here for evaluation (HL). While these do not appear in the unsupervised set, there is overlap in terms of speakers and likely content.


The text corpus of 36.6 million words used for LM training is comprised of normalised versions of the New Corpus of Ireland \cite{kilgarriff2006efficient} (c. 30m words), the Bible (c. 0.1m words), Irish language Wikipedia texts (2.9m) and the supervised training set texts (4.6m).

\subsubsection{Experiment}
The acoustic model (AM) in the baseline ASR system M\textsubscript{0} is a Time-Delay Neural Network (TDNN) \cite{peddinti2015time}, trained using the ~398h supervised train set (see Table \ref{tab:supervised_training_set}) for 4 epochs. The initial alignment is produced by a triphone GMM-HMM trained with standard MFCC features, applying linear discriminative analysis (LDA), maximum likelihood linear transformation (MLLT), feature space maximum likelihood linear regression (fMLLR) and speaker adaptive training (SAT). The features for training the TDNN model are 40-dimensional
high-resolution MFCCs stacked with 100-dimensional online extracted i-vectors. Two widely used on-the-fly data augmentation techniques for ASR – speed perturbation \cite{ko2015audio} with factors of 0.9, 1 and 1.1, and spectral augmentation \cite{park2019specaugment} were applied to augment the AM training data. The TDNN model consists of 6 TDNN layers with a hidden dimension size of 768. A pronunciation dictionary based on the Global rules, as described in \citet{qian2022test} and \citet{lonergan2022cross}, which capture cross-dialect variation in the pronunciation of phonemes and morphemes, is used, along with a 4-gram LM, trained using the text corpus described in the last paragraph of Section \ref{Data}.

As described in Section \ref{ASR}, pseudo-alignments for the unsupervised data are acquired by decoding the data using M\textsubscript{0}, rescoring the undeterminised decoding lattices and finding the best path for each utterance. Rescoring is done using a 5-gram LM trained with the same texts described in Section \ref{Data}. The resulting unsupervised and supervised alignments are combined with equal weighting. These semi-supervised alignments are then used to train M\textsubscript{1} with SpecAug for 6 epochs, with the same AM structure, lexicon and LM as M\textsubscript{0}.

Recurrent neural network LMs (RNNLM) are beneficial in rescoring n-best lists generated by an ASR system \cite{xu2018pruned}. An RNNLM is trained on the text corpora listed in Section \ref{Data} and is used in these experiments. Where results including RNNLM are reported, they are labelled as (+LM). 

\subsubsection{Results}
The WERs for in-domain test sets MG and SS have a relative improvement of 9\% and 2\% respectively with M\textsubscript{1}. For the out-of-domain test sets CV and FL, the performance improves by 14\% and 7\% relatively. For HL, which more closely matches the unsupervised data, there is a more dramatic relative performance improvement of 27\%. RNNLM rescoring improves performance across the board and is complementary with the improved, semi-supervised acoustic model. Table \ref{tab:mg_test_performance_breakdown} provides a breakdown of the performance on the MG set by speaker variety. The starkest improvement brought by the SSL approach to MG is the relative WER reduction of 17\% for Ul speakers. 

From these results, it is clear that SSL most significantly impacts performance on out-of-domain datasets, or domains more similar to the unsupervised training set (i.e. HL). Another noteworthy result is the boost in performance of Ul speakers, which is the least represented of the three dialects in the supervised training set (see Table \ref{tab:supervised_training_set}). Previous studies on Irish dialect bias in ASR have shown that the Mu and Co dialects reinforce each other in terms of performance, whereas Ul, being a more distant dialect, is an outlier \cite{lonergan2023balance}. The improvement can be explained by Ul being well represented in the unsupervised set, indicating that such dialect bias can be alleviated using SSL. The improvements could be increased further by repeating this experiment multiple times, using the student of a previous experiment as the teacher for the next, or by increasing the size of the unlabelled dataset.

\begin{table}
\begin{tabular}{l|lllll}
           & MG   & SS   & CV   & FL   & HL   \\ \hline
M\textsubscript{0} & 14.1 & 27.3 & 27.5 & 51.9 & 22.1 \\ \hline
M\textsubscript{1}    & 12.8 & 26.7 & 23.7 & 48.5 & 16.1 \\
+LM        & 10.9 & 24.0 & 19.6 & 44.5 & 14.1
\end{tabular}
\caption{ASR performance breakdown of models M\textsubscript{0} and M\textsubscript{1} of test sets and RNNLM rescoring (+LM).}
\label{tab:test_performance_breakdown}
\end{table}


\begin{table}
\begin{tabular}{l|lllll}
                & Overall & Ul   & Co   & Mu   & Nn   \\ \hline
M\textsubscript{0}      & 14.1    & 18.5 & 14.0 & 10.8 & 13.2 \\ \hline
M\textsubscript{1} & 12.8    & 15.3 & 13.1 & 10.2 & 12.6 \\
+LM             & 10.9    & 12.7 & 11.6 & 8.8 & 10.4
\end{tabular}
\caption{ASR performance breakdown by speaker variety of MíleGlór test set.}
\label{tab:mg_test_performance_breakdown}
\end{table}




\subsection{Capitalisation and Punctuation Restoration} \label{C&P}

 In this work, we propose a novel approach to tackle the punctuation and capitalisation task, namely, a sequence-to-sequence (S2S) approach which will target all the rich transcription features in an unified way. The input for such a model is uncapitalised and unpunctuated i.e., as close as possible to the actual output of the ASR system. The output is the same text in its rich transcription format with correct capitalisation and punctuation, while also including digits and acronyms. The conventional approach is to use a classifier, which for each word in the input text, predicts whether the word should be followed by punctuation or should be capitalised, however as the input and output texts do not have a one-to-one word correspondence, a S2S architecture is more appropriate.

To that aim, our proposed model is a transformer based machine translation model, implementing the original model by \citet{vaswani2023attentionneed}, which is based on attention mechanisms. We used the MarianNMT implementation \cite{mariannmt} of this architecture.

For comparison purposes we have also tested a baseline system using a classification model (CLAS), Nvidia’s Nemo Punctuation and Capitalisation Model\footnote{\url{https://docs.nvidia.com/nemo-framework/user-guide/latest/nemotoolkit/nlp/punctuation\_and\_capitalization.html}}. This model features two token-level classifiers on top of a pre-trained BERT LM. For each word in the input text, the model predicts a punctuation mark that should follow the word, if any, and predicts also if the word should be capitalized or not. The output text is then regenerated applying the classification results to each input word. The classes of the original capitalisation model are expanded to include two additional classes for 2nd and 3rd letter capitalisation. The punctuation classifier has been trained with seven classes: commas, periods, question marks, exclamation marks, semicolons, colons and none.
\subsubsection{Data}
A text corpus of 5 million Irish sentences has been used to train the model. This corpus consists of the Irish section of the Paracrawl corpus (PC\_ga) \cite{espla-etal-2019-paracrawl}, and the already mentioned New Corpus of Ireland (NCE), the Bible (BI) and the spontaneous speech corpus texts from the supervised training set, excluding the sentences used for testing (SS). The details are shown in table \ref{tab:DBtraining}. 

Four additional datasets have been used for evaluation: The Irish Language part of the FLoRes evaluation dataset (FO) \cite{nllb2022}, commonly used for machine translation evaluation for low resourced languages and the MiléGlór (MG), Fleurs-R (FL) and CommonVoice (CV) evaluation datasets, employed also for the evaluation of the ASR. The details of these databases are summarised in Table \ref{tab:DBeval} in the Appendix.

The original text corpora were cleaned to create the training and evaluation datasets, removing non-standard characters, brackets, curly brackets and parenthesis, and standardising the use of spaces, quotes and so on. This clean text is the ground truth that will be used as target dataset in the case of our machine translation model. It will be referred as rich transcription (RT) dataset. 

In order to obtain an input text as similar as possible to plain text output of an ASR system, the ground truth target was processed by the normalisation module of the Abair \cite{murphy23_sigul} speech synthesis system. The normaliser converts every digit, acronym and some symbols into pronounceable texts, keeping the punctuation and capitalisation of the text. We will refer to this dataset as normalised rich transcript (NR). This dataset is used as ground truth to train the classifier system. The normalised text is then stripped out from any non-alphabetic character and lower cased, obtaining the input dataset (IN) for both of the systems.

\subsubsection{Experiment Set-up}
The proposed S2S model has a transformer architecture, with 8 heads, 6 encoding and 6 decoding layers, transformer dropout of 0.1 and tied embeddings. The training was done using label smoothing of 0.1, learning rate of \(3\cdot10^{-4}\), warm-up stage and early stopping using cross-entropy, perplexity, BLEU detok, and CE-mean-words as validation metrics and a beam size of 6.

The baseline classifier system used the standard architecture of the NeMo model. It was trained using Google's pretrained BERT-base-uncased\footnote{\url{https://huggingface.co/google-bert/bert-base-uncased}}.

As the approach is a sequence to sequence task, to evaluate the systems, our main metric is a modified word error rate that uses the rich transcript as ground truth (instead the usual uncapitalised, unpunctuated text). We will denote it as WER\textsubscript{pc} to distinguish it from the usual WER in ASR. We are also using character error rate (CER) to mean calculated with the rich transcript as target. Along with these metrics, we also use the BLEU score \cite{post-2018-call}, a common machine-translation metric which measures the similarity between generated and reference translations using n-grams. 

\subsubsection{Experiments and Results}
We have performed two experiments to evaluate the proposed model using two metrics. Firstly, we compare the S2S approach with the baseline classifier approach. Due to the more limited capabilities of the classifier, and to allow a fair comparison of the performance of both models, we have used the normalised rich transcripts (NR) datasets as targets and the lower-cased, punctuation removed versions as input (IN). In this setup the input and output text are exactly the same with the only difference of punctuation marks and capitalisation. 

The main metric used for comparison here is accuracy. This gives a general idea of the performance of the systems although the classes are severely unbalanced. The results in Table \ref{tab:Exp1Res} show the accuracy and resulting WER\textsubscript{pc} of both classifiers: capitalisation and punctuation. Both systems perform well with the proposed S2S system showing slightly higher accuracy and better WER\textsubscript{pc}. 

\begin{table}
    \centering
    \begin{tabular}{p{0.6cm}|p{0.6cm}p{0.6cm}p{0.9cm}|p{0.6cm}p{0.6cm}p{0.9cm}}
        &\multicolumn{3}{c}{S2S} &\multicolumn{3}{c}{CLAS} \\ \hline
        &Capt  &Punct&WER\textsubscript{pc}  &Capt &Punct&WER\textsubscript{pc}\\ \hline
    FO	&0.98&0.96& 7.87&0.97	&0.96&7.93\\ \hline
    MG 	&0.98&0.95&8.36&0.97	&0.95&9.69\\ \hline
    CV	&0.97&0.89&15.17&0.96	&0.89&16.68\\ \hline
    FL	&0.97&0.96&8.27&0.97	&0.96&8.53\\ \hline
    ALL &0.98&0.95&8.40&0.97	&0.95& 9.38\\ \hline
    \end{tabular}
    \caption{Capitalisation and punctuation accuracy and WER\textsubscript{pc} using normalised rich transcripts (NR) as target.}
    \label{tab:Exp1Res}
\end{table}


The second experiment setup reflects the actual use case of the restoration system: plain text at the input (the IN dataset) and full rich transcription (RT) at the output. Table \ref{tab:EvalRes} presents 3 groups of results: No C\&P correspond to the comparison between the input (IN) and target outputs (RT) without any C\&PR system and gives an idea of the disparity of both datasets, defining the maximum error level (or minimum BLEU) that will be corrected by the restoration systems. S2S and CLAS groups correspond to the results of both systems.

\begin{table*}[t]
    \centering
    \begin{tabular}{c|ccc|ccc|ccc}
        &\multicolumn{3}{c}{No C\&PR} &\multicolumn{3}{c}{S2S} & \multicolumn{3}{c}{CLAS}\\ \hline
        &WER\textsubscript{pc}  &CER  &BLEU &WER\textsubscript{pc}  &CER  &BLEU &WER\textsubscript{pc}  &CER  &BLEU \\ \hline
    FO	&22.2	&7.7   &64.7 &7.9	&1.9	&85.1 &13.62&	5.31&	80.1\\ \hline
    MG 	&18.5	&5.0	&66.1	&8.34	&1.69	&88.4&10.18&	2.25&	84.7\\ \hline
    CV	&25.6	&5.7	&61.4 &18.29	&4.94	&82.9 &16.78&	3.19&	79.9\\ \hline
    FL	&21.7    &7.2	&64.1	&8.52	&1.99	&84.3&13.33&	4.83&	80.3\\ \hline
    ALL &19.7	&5.8	&65.5	&8.50	&1.83	&87.4&11.29&	3.14&	83.6 \\ \hline
    \end{tabular}
    \caption{C\&PR performance with rich transcription (RT) as target.}
    \label{tab:EvalRes}
\end{table*}

\begin{table*}[t]
    \centering
    \begin{tabular}{c|c|ccc|ccc}
        &ASR&\multicolumn{3}{c}{No C\&PR} &\multicolumn{3}{c}{S2S} \\ \hline
        &WER &WER\textsubscript{pc}  &CER  &BLEU &WER\textsubscript{pc}  &CER  &BLEU  \\ \hline
    MG	&10.9&26.11    & 9.65	&55.2	&18.78	&6.76	&72.3\\ \hline
    CV	&19.6&40.43	&16.59	&45.5 &34.75	&15.63	&63.0\\ \hline
    FL 	&44.5&54.58	&30.35	&24.9	&50.70	&28.65	&31.9\\ \hline
    \end{tabular}
    \caption{Performance of the S2S C\&PR system on ASR generated text with rich transcriptions (RT) as target.}
    \label{tab:EvalASR+CapPunct}
\end{table*}

The results show that both systems are effective in C\&PR, obtaining important reductions in the WER\textsubscript{pc} and CER metrics. Our proposed S2S system reduces the WER\textsubscript{pc} and CER by more than 50\% and improves the BLEU more than 20 points for all datasets. S2S clearly outperforms the baseline classifier in this experiment, because it not only restores the casing and punctuation more effectively, but also changes digits or acronyms to a textual output that is closer to the target rich transcription. The error rates for S2S are below 9\% for all the databases, except CV. This database contains a large number of very short, fragmentary sentences with inconsistent casing and punctuation, which may be interpreted as titles.

Finally Table \ref{tab:EvalASR+CapPunct} shows the effect of applying the S2S system to the actual output of the ASR. The input text is the plain text generated by the M\textsubscript{1}(+LM) ASR model. The WER of these texts compared to the uncapitalised and unpunctuated references as shown previously in Table \ref{tab:test_performance_breakdown} is shown in the column ASR for readability. The results in column No C\&PR show the WER\textsubscript{pc} of the same texts when they are compared to the rich transcription. The columns under S2S show the results when the restoration system is applied. The final WERs are always lower than the accumulated WERs of the ASR and the S2S, suggesting that the degradation in the input text generated by the ASR does not impact in the performance of the S2S system.

\section{Conclusion and Future Work}

This paper constitutes an important step towards democratising speech-related AI technologies for the Irish language and its speakers. The ASR experiments have demonstrated that SSL learning is an attractive solution to improve performance for out-of-domain datasets and underrepresented dialects in the supervised training set. As stated, this improvement can be increased by iteratively repeating this process or increasing the size of the unlabelled dataset. Future work will explore SSL further.

The S2S model offers an elegant solution to the C\&PR problem, improving significantly over the baseline due to its ability to effectively deal with the lack of a one-to-one relationship between the outputs of an ASR system and the rich transcriptions. The S2S model could additionally be trained using the output texts of the ASR systems so that it will be able to correct of some of the ASR errors. Furthermore, it can be trained to convert specific keywords, such as punctuation symbol names, currency or units, which would be very useful for dictation applications.

\section*{Acknowledgments}
This work is part of the ABAIR initiative, which is supported by the Department of Tourism, Culture, Arts, the Gaeltacht, Sport and Media, with funding from the National Lottery, as part of the 20-year Strategy for Irish. The publication was partly funded by the Irish Research Council under grant
number 214047. Part of this work was supported by the José Castillejo Mobility Programme of the Spanish Ministry of Science, Innovation, and Universities (CAS23/00294). The C\&PR module is partially based on findings from the HiTZketan project of the UPV/EHU (COLAB22/13). We would like to especially acknowledge Dr. Gorka Labaka for his collaboration in the initial implementation of this system.

\bibliography{coling_latex}

\appendix

\section{Appendix: Additional Tables}
\label{sec:appendix}

\begin{table}[H]
\begin{tabular}{l|lllll}
                              & Full  & Ul   & Co    & Mu    & Nn   \\ \hline
Syn                   & 41.4  & 22.2 & 8.8   & 10.4  & -    \\
MG                      & 17.3  & 3.1  & 5.6   & 3.4   & 5.2  \\
SS            & 259.7 & 55.5 & 104.9 & 95.4  & 3.9  \\
AB                    & 36.6  & 10.3 & 2.5   & 14.6  & 9.2  \\
CCh               & 25    & -    & 25.0  & -     & -    \\
BBhÁ & 17.9  & -    & 17.9  & -     & -   \\ \hline
\textbf{Total}                         & \textbf{397.9} & \textbf{91.1} & \textbf{164.7} & \textbf{123.8} & \textbf{18.3} 
\end{tabular}
\caption{Duration breakdown in hours of ASR training set by speaker variety.}
\label{tab:supervised_training_set}
\end{table}

\begin{table}[H]
\begin{tabular}{l|lllll}
            & Total Dur (h)& Ul & Co & Mu & Nn \\ \hline
MG    & 10.2      & 2.5    & 3.4    & 3.0    & 1.3    \\
SS & 28.2      & 7.5    & 10.4   & 10.3   & -     
\end{tabular}
\caption{Duration breakdown by speaker variety of MíleGlór and Spontaneous Speech test sets.}
\label{tab:test_set_by_speaker_var}
\end{table}

\begin{table}[H]
\begin{tabular}{l|ll}
               & Dialect & Dur (h)       \\ \hline
Adhmhaidin     & Co & 779.0           \\
Barrscéalta    & Ul & 1002.4          \\
Saol Ó Dheas   & Mu & 993.9           \\
Nuacht         & mix &  400.0           \\ \hline
\textbf{Total} & - &\textbf{3230.0}
\end{tabular}
\caption{Duration in hours and dialect information of unsupervised set by radio show}
\label{table:unsupervised_speech_corpus}
\end{table}

\begin{table}[H]
\begin{tabular}{l|lllll}
         & MG    & SS     & CV  & FL  & HL  \\ \hline
\#Utts     & 8,423 & 19,266 & 516 & 548 & 198 \\
Dur (h) & 10.2 & 28.2  &  0.6  &  2.2   & 0.7    
\end{tabular}
\caption{Number of utterances and total duration of test sets.}
\label{tab:test_set_duration_num_utts}
\end{table}

\begin{table}[H]
    \centering
    \begin{tabular}{c|ccc}
        &\#lines  &\#words  &\#chars \\ \hline
    FO	&1012	&25772	&163254 \\ \hline
    CV	&515	&3423	&19617\\ \hline
    FL	&548	&13634	&86282\\ \hline
    MG 	&8423	&98463	&559675 \\ \hline
    \end{tabular}
    \caption{Features of the databases used for evaluation}
    \label{tab:DBeval}
\end{table}

\begin{table}[H]
    \centering
    \begin{tabular}{c|ccc}
         &\#lines  &\#words  &\#chars \\ \hline
    PC\_ga     & 3.2&63.1&417.2\\ \hline
         NCE&1.8&30.4&181.5  \\ \hline
         BI& 0.3&	0.8&	4.5 \\ \hline
         SS&1.4&3.6&20.6 \\ \hline
    \end{tabular}
    \caption{Features of the databases used for training the Cap\&Punct systems (numbers in millions)}
    \label{tab:DBtraining}
\end{table}

\end{document}